# E-TRoll: Tactile Sensing and Classification via A Simple Robotic Gripper for Extended Rolling Manipulations*

Xin Zhou, *Student Member, IEEE*, and Adam J. Spiers, *Member, IEEE*

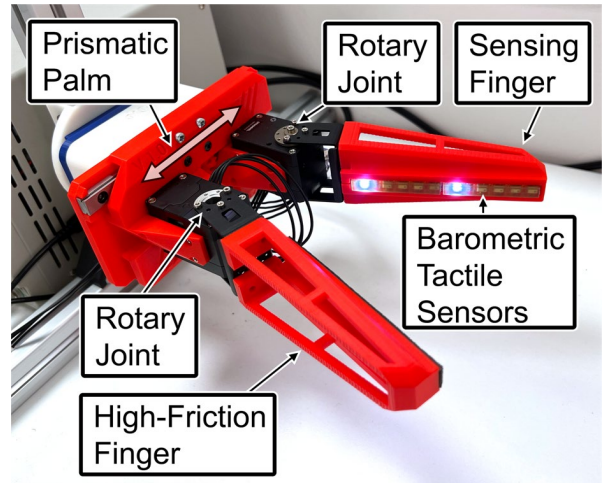

Figure 1: The robotic hand consists of a prismatic variable-width palm, two fingers with rotary joints, and a barometric tactile sensor array.

*Abstract*—Robotic tactile sensing provides a method of recognizing objects and their properties where vision fails. Prior work on tactile perception in robotic manipulation has frequently focused on exploratory procedures (EPs). However, the also-human-inspired technique of in-hand-manipulation can glean rich data in a fraction of the time of EPs. We propose a simple 3-DOF robotic hand design, optimized for object rolling tasks via a variable-width palm and associated control system. This system dynamically adjusts the distance between the finger bases in response to object behavior. Compared to fixed finger bases, this technique significantly increases the area of the object that is exposed to finger-mounted tactile arrays during a single rolling motion (an increase of over 60% was observed for a cylinder with a 30-millimeter diameter). In addition, this paper presents a feature extraction algorithm for the collected spatiotemporal dataset, which focuses on object corner identification, analysis, and compact representation. This technique drastically reduces the dimensionality of each data sample from 10×1500 time series data to 80 features, which was further reduced by Principal Component Analysis (PCA) to 22 components. An ensemble subspace k-nearest neighbors (KNN) classification model was trained with 90 observations on rolling three different geometric objects, resulting in a three-fold cross-validation accuracy of 95.6% for object shape recognition.

## I. Introduction

Although humans substantially depend on vision to perform everyday tasks, the sense of touch also plays an essential role. Not only do we rely on touch when vision is not an option, such as when we are looking for house keys in trouser pockets or locating the light switch in a dark room, we also use touch to sense finer shapes and material properties that are often difficult or impossible to determine through the naked eye [1]. The field of robotics has long recognized the practicality of our somatosensory system, and in recent decades has been rapidly developing tactile sensors based on a wide range of technologies [2]–[4], to perform tasks such as texture recognition [5], [6] stiffness measurement [6], slip detection [7], [8], and grasping force feedback [8].

A less-commonly researched haptic property is determining features of object contour or shape. One approach to this problem is to utilize a high-density tactile array to consecutively extract the local shape of small areas of an object's surface. This approach of using exploratory procedures (EPs) was taken in 1993 [9] as well as in recent years [10], [11]. Two fallbacks are noticeable in those pieces of work: Firstly, the need for the object to be immovable. Physical contact often impacts the position and orientation of an object, which increases the difficulty of recognizing the object's overall shape [12]. Although tactile data processing methods have been proposed to compensate for this issue [12], researchers often still simply use glue or clamps to fix the object to a desk or other supports [10]. Secondly, this approach requires precise and repeated contact, which is time-consuming. For example, in [11], a full object exploration consisting of 5 EPs takes around 85 seconds.

These two issues give rise to a different approach for tactile data gathering – robotic In-Hand Manipulation (IHM). IHM refers to "the task of changing the grasp on a hand-held object without placing it back and picking it up again" [13]. In our daily life, these tasks are performed frequently and usually without much thought, such as aligning a key with its lock or writing a letter with a pen. When we attempt to gauge the shape of an object through touch, we often pick it up and use IHM to expose the surface contour of the object to the mechanoreceptors of the fingertips [1].

The tactile data collection method used by our robotic hand also follows an IHM approach. A low-cost ($300) tactile array is mounted on one of the two fingers. Through rolling the object between the two fingers, the tactile array is exposed to a larger portion of the object's surface than would be achieved by grasping alone. This is similar to rolling a small object between the thumb and the index finger on a human hand to estimate its shape. This approach of tactile rolling was recently explored by Mohtasham *et al.* [14], showing promising results for object recognition. In that work, the robot hand had a similar design, consisting of two rotary fingers but with a fixed palm width. We made two main improvements to this previous approach:

* Research supported by Imperial College London internal funds.
Both authors are with Imperial College London (corresponding author: xin.zhou16@imperial.ac.uk).

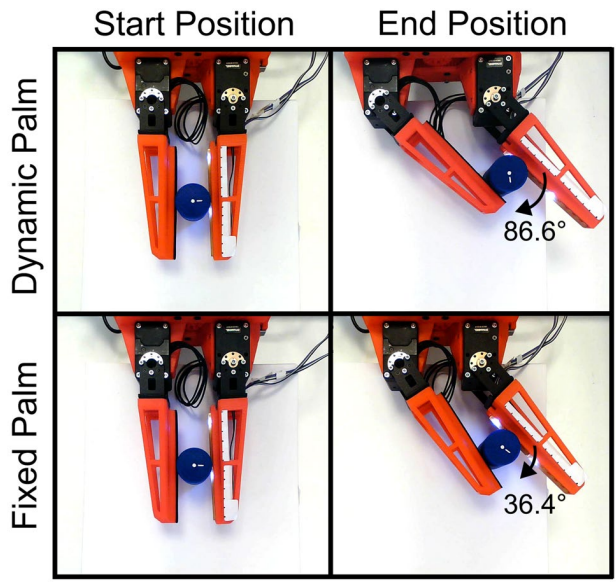

Figure 2: Rolling a cylinder with 30 mm diameter with a dynamic palm versus a fixed palm width of 68.5 mm. An increase in object rotation of 137.9% is observed.

*1) Extended rolling range*

The manipulability of a large range of two-finger robotic hands were extensively investigated and compared by Bircher *et al.* in 2021 [15]. Their work inspired us to add the prismatic palm to Mohtasham *et al.*'s 2DOF hand design of fixed rotary joints at the finger bases. We found that by utilizing a 1DOF prismatic palm, the object contact area during a single rolling exploration can be significantly increased (Fig. 2). This is achieved by dynamically varying the distance between the two finger bases with the goal of keeping the fingers parallel to each other throughout manipulation.

*2) Generalizable data processing approach*

Mohtasham *et al.*'s shape recognition algorithm was prepared as a proof of concept for the available object rolling data and is highly fitted for that work's choice of objects. The algorithm mainly consists of hand-designed conditional-statements that do not generalize well. In this paper, we have designed the basis of a tactile feature extraction algorithm intended to be applicable to all rollable objects with faces and corners, which the algorithm is able to distinguish. Though this is currently only tested with geometric primitives, our intention is to extend this algorithm in the near future to also deal with irregular objects. Currently, an ensemble of Random Subspace and K-Nearest Neighbors is used to recognize object shapes based on these features.

Two more pieces of work are noteworthy in the context of fast tactile object recognition via robotic hands. Firstly, variable-friction fingers mounted on rotary joints have previously been used to achieve object identification via rolling and sliding IHM [16]. Their work did not require tactile sensors, solely using proprioceptive information from the servo motors to successfully classify objects via an Extra Trees classifier [17]. Our work provides a simpler mechanical finger design using fewer actuators and a faster IHM procedure, due to the eliminated need to switch between rolling and sliding finger surfaces. Secondly, Spiers *et al.* previously used underactuated two-link fingers mounted with tactile sensors to identify objects through a single grasp [18]. Prior to machine learning, both works decompose time-series data into a compact set of numerical features. However, those features were not directly related to real-world object properties, as in our new approach, in which they are linked to faces and corners. We feel that this identification of tangible characteristics provides a more generalizable and scalable framework.

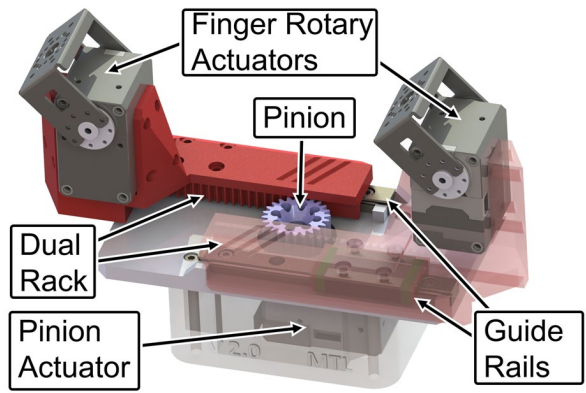

Figure 3: The prismatic palm is implemented through a dual rack and pinion mechanism. Both the sensor finger and the high friction finger are directly connected to the servo horns, forming rotary joints.

## II. System Overview

We have developed a robotic hand optimized for collecting tactile data whilst rolling prismatic objects. We have named this hand E-TRoll (Extended Tactile Rolling) due to its capability of extending the rolling range via the dynamically controlled prismatic palm. This section provides details on the mechanical design and control systems that make up the robotic hand shown in Fig. 1.

### A. Mechanical Design

The gripper consists of two 1DOF fingers of length 132 millimeters, with a prismatic palm that can adjust the distance between the two finger joints from 50 to 150 millimeters. The dual rack and pinion design of the adjustable palm is inspired by Elangovan et al.'s adaptive gripper [19] and Spiers' 'Haptic Taco' shape changing haptic interface [20]. Fig. 3 presents a 3D rendered view of the adjustable palm mechanism. Most structural elements of the gripper are 3D printed in polylactic acid (PLA) on a Raise3D E2 printer, with the exception of the two MGN9H linear guide rails manufactured by Yanmis. The rails are of size 100mm × 9mm and are used to ensure a stable prismatic movement, free of the binding effects of 3D printed guide structures.

One of the fingers acts as the sensing finger and holds two TakkStrip 2 barometric tactile sensor arrays previously manufactured by RightHand Robotics [21] based upon the work described in [2]. These sensor arrays are no longer supported or manufactured. Each of the two TakkStrips consists of six Bosch MEMs barometric sensors, arranged in a single line on a PCB, adding up to twelve barometric sensors on the sensing finger. The other finger provides a high friction surface (a section of a rubber mouse pad) to aid the rolling task

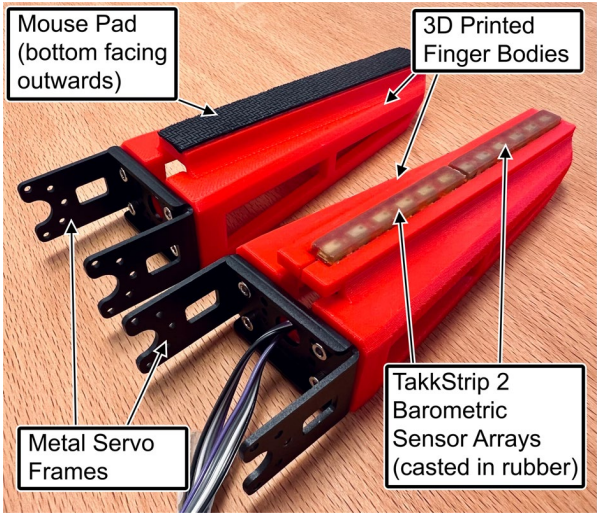

Figure 4: A high-friction finger (left) and one sensing finger (right) are used by E-TRoll.

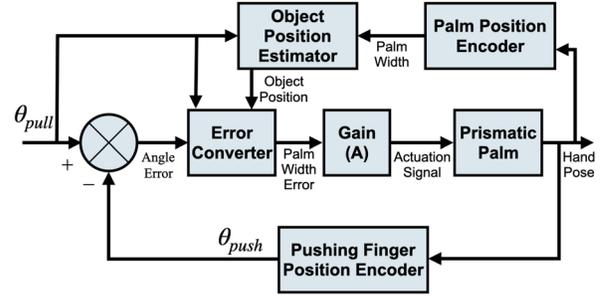

Figure 6: Block diagram for the Prismatic Palm Controller.

(Fig. 4). Three Dynamixel XM430-W350-R servo motors are used for the prismatic and rotary joints.

The advantage of using the prismatic palm mechanism is significant. To demonstrate, a 30-millimeter diameter cylinder was rolled between the fingers with and without dynamic palm width (Fig. 2). The right finger was instructed to start perpendicular to the base and stop after rotating 44°. With a fixed palm width (as in Mohtasham's work [15]) of 68.5 mm, the object only rotated by 36.4°. Using the prismatic palm to keep the two fingers parallel, an object rotation of 86.6° was achieved. This translates to 34.2 mm of the cylinder's base circumference touching the tactile array using a dynamic palm, compared to only 21.0 mm with a fixed palm—an increase of 62.9%.

### B. Control Systems

#### 1) Rotary Joint Control

Our control approach is based upon the technique previously discussed in [16]. Note that compared to [16], our torque controller is subject to less noise and disturbance, due to a direct mechanical coupling between the actuator and finger via an aluminum bracket, manufactured by Robotis (Fig. 4). The variable friction gripper on the other hand utilized an OpenHand base, whose transmission relied on tendon routing across multiple (metal and 3D-printed) surfaces and a return spring which opposed inwards motion.

The Dynamixel model-XM servos were set to operate in position control and current control modes. The latter attempts to keeps the current (and thus torque) of the servo to a specified value. To ensure a firm grasp of the object during in-hand-manipulation, the finger joint actuators follow a "push and pull" approach: The pulling finger is in position control mode and slowly rotates away from the pushing finger, whilst the pushing finger is in current control mode to apply a controlled force on the object to press it against the pulling finger. In the example provided in Fig. 5, the right finger is the pulling finger, and the left finger is the pushing finger. When the object is rolled in the opposite direction, the roles of the fingers are reversed.

#### 2) Prismatic Palm Controller

The goal of the palm controller is to ensure that the two fingers are parallel to each other throughout the rolling motions. Fig. 5 demonstrates the rolling of a cylindrical object—note the changing distance between the two finger bases at different steps of the rolling task.

Fig. 6 shows the block diagram for the implemented prismatic palm control system. A controller with negative feedback is chosen for this task, with the goal to minimize the difference between the two finger joint angles.

After testing a simple control system with a gain proportional to the angle error between the two fingers, we found that the system tends to oscillate when the fingers are near perpendicular to the base. This is caused by the palm width error not being proportional to the finger angle error—at different pulling finger angles and object positions, the palm needs to adjust for a significantly different amount to compensate for the same angle error. This raises the need for the 'Error Converter' block to calculate the palm width error.

The block labelled 'Object Position Estimator' uses the palm width and pulling finger angle to approximate the object position. The result, in conjunction with the pulling finger angle, converts the errors via the Error Converter. Both the Error Convertor and Object Position Estimator will now be explained further.

##### a) Error Converter

In Fig. 7, $w$ denotes the current palm width and $dw$ denotes the palm width error. The angles of the pushing and pulling finger are $\theta_{push}$ and $\theta_{pull}$, whilst $l$ and $l_{mid}$ refer to the distances along the pushing finger, from base to contact point and finger midpoint respectively. $\theta_{Target}$ refers to the angle of the target pushing finger and the angle between pushing finger and its target is labelled as $-d\theta$. The negative sign is due to the larger pushing finger angle compared to the target angle. The target pushing finger is parallel to the pulling finger, resulting in an equal angle ($\theta_{Target} = \theta_{Pull}$). According to the sine rule:

$$\frac{l}{\sin(\theta_{Pull})} = \frac{dw}{\sin(-d\theta)} . \quad (1)$$

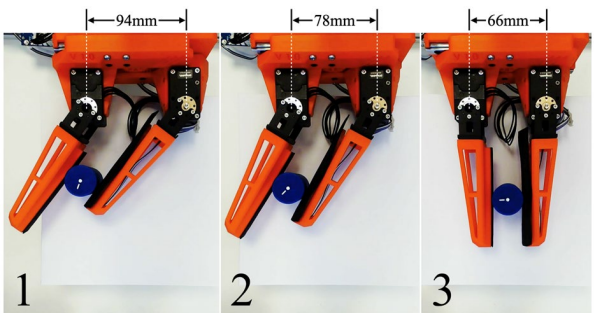

Figure 5: An example rolling manipulation with the prismatic palm controller enabled. Whilst the fingers rotate from an angled position (step 1) to a straight position (step 3), the controller adjusts the distance between the finger bases to keep the fingers parallel.

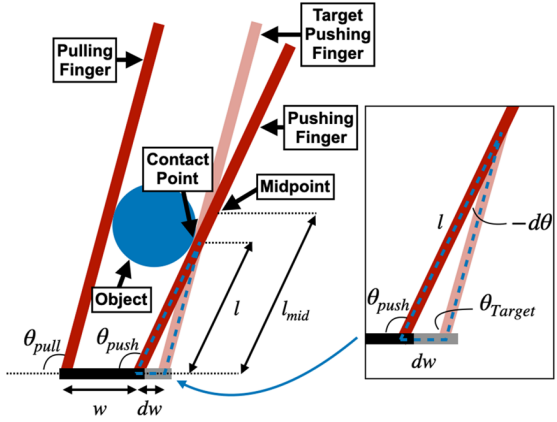

Figure 7: The palm width error ($dw$) depends on pulling finger angle ($\theta_{Pull}$) and the object's distance to the palm along the pushing finger ($l$).

Since $d\theta$ is sufficiently small, we can approximate $\sin(-d\theta) = -d\theta$, allowing us to rearrange (1) as follows:

$$\frac{dw}{d\theta} = -\frac{l}{\sin(\theta_{Pull})} . \quad (2)$$

This relationship expresses the ratio between palm width error and angle error. Multiplying the right-hand side with the angle error gives us the desired palm width error.

*b)* *Object Position Estimator*

We specify that the object rolling procedure begins with both fingers at a 90-degree angle relative to the base. We approximate the shape of convex objects as cylinders and presume the two fingers to be parallel throughout rolling. This enables us to visually fix the center of the object and constrain the fingers to be horizontal, resulting in the configuration shown in Fig. 8. The thicker pale red lines indicate the initial positions of the fingers and palm, and the thinner black lines show the current positions. Similar to a double rack and pinion mechanism, the two fingers have been displaced by the same distance $l_{roll}$ in opposite directions compared to the initial positions. As a result, the initial and current palms cross each other at their midpoints. Observing the small, shaded triangle in the bottom left corner of the image, we can calculate $l_{roll}$:

$$l_{roll} = \frac{w}{2}\cos(\theta_{Pull}) . \quad (3)$$

Initially, the object is always placed within a 20-millimeter range centered at the midpoint. Thus, we approximate the initial contact point to overlap with the pulling finger midpoint. Thus, the final object position $l$ is expressed as:

$$l = l_{mid} - \frac{w}{2}\cos(\theta_{Pull}) . \quad (4)$$

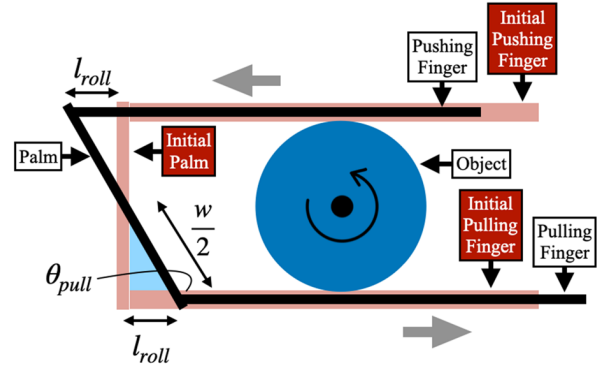

Figure 8: The object's position can be estimated by calculating the distance it has rolled from the initial position.

In practice, the Error Converter and Object Position Estimator are calculated as a single, more elegant equation combining (2) and (4):

$$\frac{dw}{d\theta} = -\frac{l_{mid} - \frac{w}{2}\cos(\theta_{Pull})}{\sin(\theta_{Pull})} ,$$

$$\frac{dw}{d\theta} = \frac{w}{2}\cot(\theta_{Pull}) - l_{mid}\csc(\theta_{Pull}) , \quad (5)$$

where $l_{mid}$ is measured to be 85mm. Despite the fact that multiple approximations were made whilst deriving the final equation, its implementation successfully eliminated the previously observed oscillations.

### III. METHODS

*A. Objects*

Three objects were 3D printed with PLA on a Raise3D E2 with 25% infill: a hexagonal prism, a circular prism (cylinder), and a square prism (cuboid). Each of these prisms have a height of 50 mm and a base with an inner diameter of 30 millimeters (Fig. 9).

*B. Sensor Calibration*

Around room temperature, each barometric sensor outputs highly linear pressure values with regards to applied normal force, until saturation [2]. A calibration rig designed by

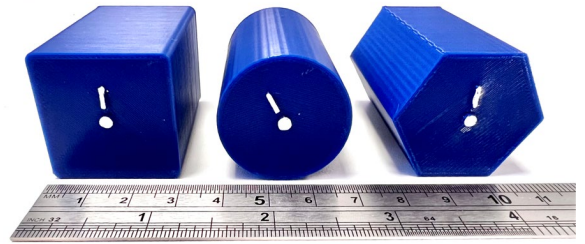

Figure 9: The three objects used in the experiments. From left to right: a square prism (cuboid), a cylinder, and a hexagonal prism.

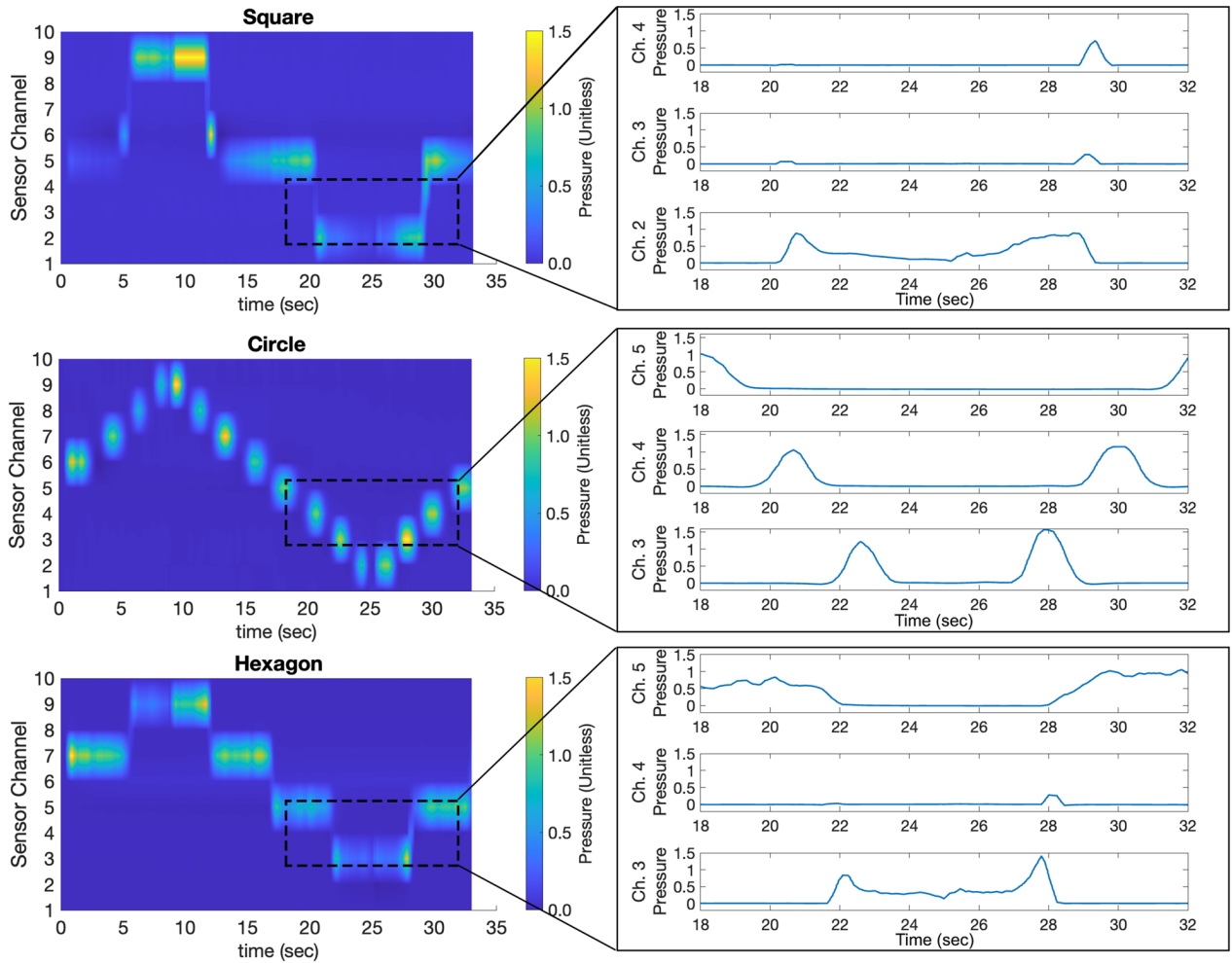

Figure 10: Heatmaps for the tactile data collected through rolling (left) and line plots for selected sensor channels between 18 and 32 seconds (right).

Mohtasham *et al.* [14] was used to align weights of 7.75 grams, 19.04 grams, and 29.70 grams with the center of each sensor. The readings were linearly fitted to calculate the calibration parameters for each sensor.

### C. Rolling Experiments

A pre-defined manipulation procedure was implemented for this experiment as outlined in Table 1.

During this procedure, the measurements of the barometric sensors were collected at a rate of 45 Hz. Due to a faulty sensor at the base of the tactile array, only the middle ten sensors were utilized. The starting positions and orientations of the objects are deliberately varied, by placing the object randomly within a range of 20 millimeters centered at the sensor array's midpoint, and at a random angle. This variation captures more information on how the sensors reacts when a corner of an object is pushed directly onto a barometric sensor versus somewhere between two sensors. This is highly necessary due to each sensor's asymmetric design, whose effect is explained later in this section.

The procedure was repeated 30 times for the cylinder, hexagonal prism, and for the cuboid, resulting in 90 spatiotemporal samples. No object sliding was noticeable during all rolling experiments. The duration of each data collection run is slightly different due to object shape and initial condition, varying from 32.6 seconds to 33.8 seconds. 1500 ± 50 data samples for each of the ten sensors were collected for each run, resulting in a dataset of dimensions 90 by 10 by 1500 ± 50.

Fig. 10 presents one collected spatiotemporal data sample each for rolling the square, circle, and hexagon shaped objects, following sensor calibration and processing the data via a moving average filter with a window size of 20 samples. The plots are presented both as temporal heatmaps (where brighter colours relate to higher sensor readings at that time instance) and line plots of the individual amplitudes of two selected sensors over time.

TABLE I. PREDEFINED ROLLING PROCEDURE

| Step | Description |
|------|-------------|
| 1 | Both fingers start perpendicular to the base with the object between tactile sensors 6 and 7. This arrangement is illustrated in Fig.5, image 3. The sensor finger is assigned as the pushing finger. |
| 2 | Rotate the pulling finger clockwise by 44 degrees. The pushing finger ensures a firm grasp of the object. |
| 3 | Swap pushing and pulling finger and rotate the new pulling finger counterclockwise by 88 degrees. |
| 4 | Swap pushing and pulling finger again and rotate the pulling finger back until perpendicular to the base. |

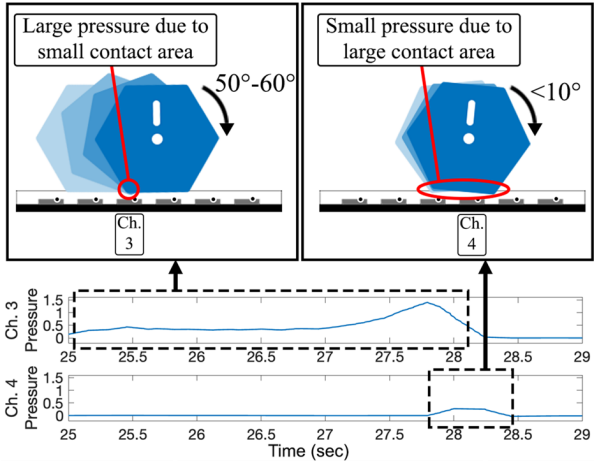

Figure 11: Pressure data as a hexagonal object rolls directly over the sensors. Sensor channel 3 is in contact with the object's edge, whereas channel 4 is in contact with a flat surface. A black dot on each barometric sensor indicates the sensor hole, which is the most sensitive region. Note that this hole is not centered, causing asymmetric sensor behavior.

*D. Feature Extraction Algorithm*

From the temporal heat plots in Fig. 10, it is clear that most of the values for each sensor channel at a specific time are zero. Further inspecting the line plots for the selected channels and time frames, we observe that the contact between the objects and the tactile array result in differently shaped peaks on each sensor channel. The goal of our feature extraction algorithm is to describe those peaks.

Fig. 11 attempts to give some insight into the structure of the rolling data collected by the barometric sensors when it encounters object corners (edges) compared to flat surfaces (faces). When rolling a polygonal prism over the tactile array, the flat surfaces only briefly meet the sensing finger, whilst the object is pivoting over an edge most of the time. Hence, an edge creates a much wider peak (sensors channel 3 in Fig. 11) compared to a flat surface (sensor channel 4 in Fig. 11). Thus, the width of a peak is an important feature to excerpt.

Flat surfaces also create less pressure when meeting a sensor as the force (provided by pressure from the opposing finger) is distributed across a larger surface area, resulting in a lower amplitude. Consequently, the amplitude of a peak is another important feature.

Comparing the peaks at sensor 3 and 5 in the same figure, it may be seen that the shapes are different, albeit both being created by an edge. Sensor 3's peak has a distinct negative skew, whereas sensor 5's peak is more symmetric. The negative skew is possibly explained by the positioning of the edge relative to the barometric sensors underneath—an edge slightly left to a sensor's center (assuming the object is rolled towards the right) would have the highest sensor reading when the next flat surface is almost reached, at which point the object briefly presses against the center of the underlying sensor. However, a symmetric peak does not necessarily entail that the edge is directly placed above the sensor center. Each

TABLE II. EXTRACTED FEATURES FOR EACH SPIKE

| Feature | Description |
| --- | --- |
| Amplitude | The maximum measured pressure during a peak. |
| Time-to-peak (TTP) | The time at the peak value, measured from the start of the predefined rolling procedure. |
| Peak width | Duration from when the value first rises above the threshold[a] (startpoint), to when it first drops below the threshold again (endpoint). |
| Skewness | Defined as the time difference between the temporal mid-point and the TTP of a peak, divided by the peak width and multiplied by a factor of 100. |

a. The threshold is defined as a measured pressure of 0.05 units, or 20% of the maximum reading within a sensor channel, whichever is larger.

Bosch barometric sensor in the TakkStrip 2 array has an asymmetric design, with the sensor hole placed closer to one end of the sensor. This means that the same force applied at the same distance to the sensor center, but on opposite sides, would result in different readings. Nevertheless, although not straightforward, the skewness of a peak (i.e. its degree of asymmetry) likely holds information on the position of edges.

Finally, the time instance of each peak's center (measured from the start of the rolling manipulation) is stored as another feature.

Table 2 summarizes the features extracted from each peak and explains how they are calculated. As our predefined rolling procedure rolls an object over each sensor twice, each sensor usually picks up a maximum of two peaks. Hence, we decided to allocate two sets of these four features to each of the ten sensors, resulting in 80 features in total. Fig. 12 presents an example of marked peak features for rolling a hexagon.

*E. Machine Learning Model*

The goal of the model is to classify the tactile data into the three object categories, using the extracted features as inputs. The Classification Learner tool, part of MATLAB's Machine Learning and Deep Learning toolbox, was used to train and compare various classification models.

Principal component analysis (PCA) is applied to the feature set before training, generating 22 components with a 95% variance. Table 3 presents the accuracies for a selection of classification models using three-fold cross validation. An ensemble of Random Subspace and K-Nearest Neighbors

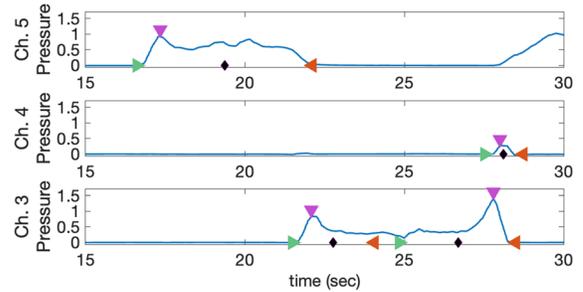

Figure 12: Sensor channels 3 to 5 of rolling a hexagon with the following peak features marked for each peak: startpoint (green right-pointing triangle ▶), endpoint (red left-pointing triangle ◀), time-to-peak and amplitude (purple down-pointing triangle ▼), and temporal mid-point (black diamond ◆).

TABLE III.  THREE-FOLD CROSS VALIDATION ACCURACIES

| Model | Validation Accuracy |
| --- | --- |
| Linear Discriminant | 62.2% |
| Gaussian Support Vector Machine (SVM) | 82.2% |
| K-Nearest Neighbors (KNN) | 70% |
| Ensemble: Bagged Trees | 75.6% |
| **Ensemble: Random Subspace with KNN** | **95.6%** |
| SVM Kernel | 76.7% |

(Subspace KNN) stands out with a validation accuracy of 95.6% and is our model of choice.

## IV. RESULTS

Fig. 13 shows the cross-validation confusion matrix of the trained Subspace KNN model. Three squares and one hexagon were wrongly predicted to be hexagons and a circle respectively. The validation accuracy of 95.6% indicates that our feature extraction method can retain enough information to recognize objects via a machine learning approach.

Comparing the surface plots of a wrongly predicted square against an example of a correctly predicted one (Fig. 14), the reason for the misclassifications is not obvious. Visually, the wrongly predicted surface plot is identifiable as a square and is similarly or even less noisy compared to the example of a correct one. However, inspecting the extracted peak features from the whole dataset, a small number of peaks are incorrectly identified for both correctly and incorrectly classified shapes. For example, the bottom sample in Fig. 14 only marked a small portion of the wide, negatively skewed peak as a peak. The top sample on the other hand, identified a wide peak as two separate peaks. The incorrectly identified peaks likely have negatively impacted the prediction as well as training of the model, making them a possible reason for the small observed validation error.

Figure 13: The validation confusion matrix of the trained Subspace KNN model via three-fold cross-validation.

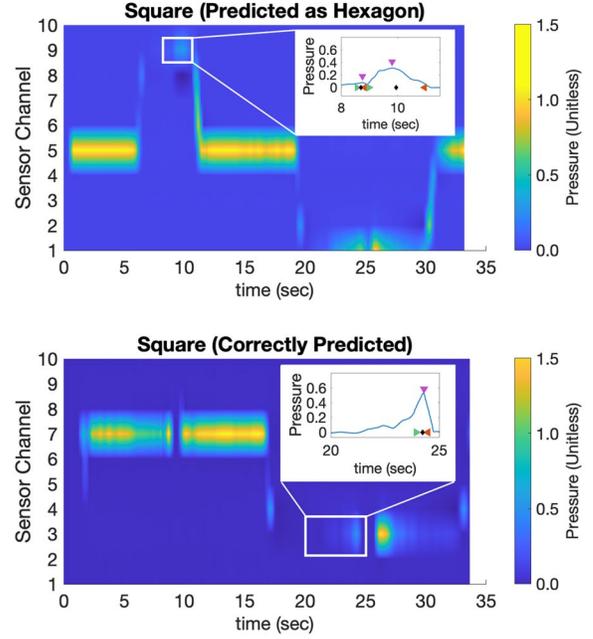

Figure 14: Comparison of a wrongly predicted and a correctly predicted square.

## V. CONCLUSIONS

The work presented has demonstrated a novel robotic gripper designed for tactile rolling tasks with extended surface object contact. In addition, a novel feature extraction algorithm was designed to capture information on object's local shapes at contacts with the tactile sensing surface (e.g., corners and flat surfaces).

The robotic hand maximizes the contact area between object and tactile array by dynamically adjusting the distance between the rotary joints at the two finger bases.

The feature extraction algorithm significantly decreases the data dimensions of the collected tactile data. The extracted features retain enough information to successfully perform object shape classification via an ensemble of Random Subspace and K-Nearest Neighbors machine learning models.

This work acts as a proof-of-concept for the proposed robotic hand and feature extraction algorithm, and consequently has a large scope for improvement. Firstly, the peak detection algorithm occasionally misidentifies start- and endpoints and needs to be improved upon. Secondly, the pre-defined rolling procedure currently rolls an object across the tactile array twice. It should be investigated if a single rolling motion can achieve similar results, which would more than half the rolling time and half the required feature set.

Finally, the extremely high classification accuracy indicates that more difficult machine learning tasks are possible. Such tasks include increasing the number of possible shape categories, and dealing with irregular objects and object pose estimation. Furthermore, we believe that extensions of this algorithm would allow further determination of corner and face characteristics, such as angle and spacing. This could pave the way for our longer-term goal of tactile shape reconstruction via IHM.